\documentclass[letterpaper, 10 pt, conference]{ieeeconf}  

\IEEEoverridecommandlockouts                              

\usepackage{cite}
\usepackage{amsmath,amssymb,amsfonts}
\usepackage{graphicx}
\usepackage{textcomp}
\usepackage{xcolor}
\usepackage{hyperref}
\usepackage{caption}
\usepackage{subcaption}
\usepackage{booktabs}
\usepackage{algorithm}

\usepackage{algorithm}
\usepackage{algpseudocode}
\usepackage{xspace}
\usepackage{lineno}
\usepackage{adjustbox} 

\newcommand{\M}{\text{UrbanHuRo}\xspace}
\newcommand{\UM}{\text{KSubMR}\xspace}
\newcommand{\DM}{\text{DSRQN}\xspace}

\newcommand{\fd}{{FastD}\xspace}
\newcommand{\bs}{{HighS}\xspace}
\newcommand{\jd}{{JointDS}\xspace}

\title{\LARGE \bf
UrbanHuRo: A Two-Layer Human-Robot Collaboration Framework \\for the Joint Optimization of Heterogeneous Urban Services
}

\author{Tonmoy Dey$^{1\dag}$, Lin Jiang$^{1\dag}$, Zheng Dong$^{2}$, Guang Wang$^{1*}$%
\thanks{$^\dag$These authors contributed equally (co-first authors).}%
\thanks{$^*$Corresponding author.}%
\thanks{$^{1}$Department of Computer Science, Florida State University.}%
\thanks{$^{2}$Department of Computer Science, Wayne State University.}%
}

\begin{document}

\maketitle
\thispagestyle{empty}
\pagestyle{empty}

\begin{abstract}
In the vision of smart cities, technologies are being developed to enhance the efficiency of urban services and improve residents’ quality of life. However, most existing research focuses on optimizing individual services in isolation, without adequately considering the reciprocal interactions among heterogeneous urban services that could yield higher efficiency and improved resource utilization. For example, human couriers could collect traffic and air quality data along their delivery routes, while sensing robots could assist with on-demand delivery during peak hours, enhancing both sensing coverage and delivery efficiency. However, the joint optimization of different urban services is challenging due to their potentially conflicting objectives and real-time coordination in dynamic environments. In this paper, we propose \M, a two-layer human-robot collaboration framework for joint optimization of heterogeneous urban services, demonstrated through the examples of crowdsourced delivery and urban sensing. There are two innovative designs in \M, i.e., (i) a scalable distributed MapReduce-based K-Submodular maximization module for efficient order dispatch and (ii) a deep submodular reward reinforcement learning algorithm for sensing route planning. Experimental evaluations on real-world datasets from a food delivery platform demonstrate that our \M improves sensing coverage by 29.7\% and courier income by 39.2\% on average in most settings, while also significantly reducing the number of overdue orders.
\end{abstract}

\section{Introduction}
\begin{figure}[t]
\centering
\includegraphics[width=0.47\textwidth, keepaspectratio]{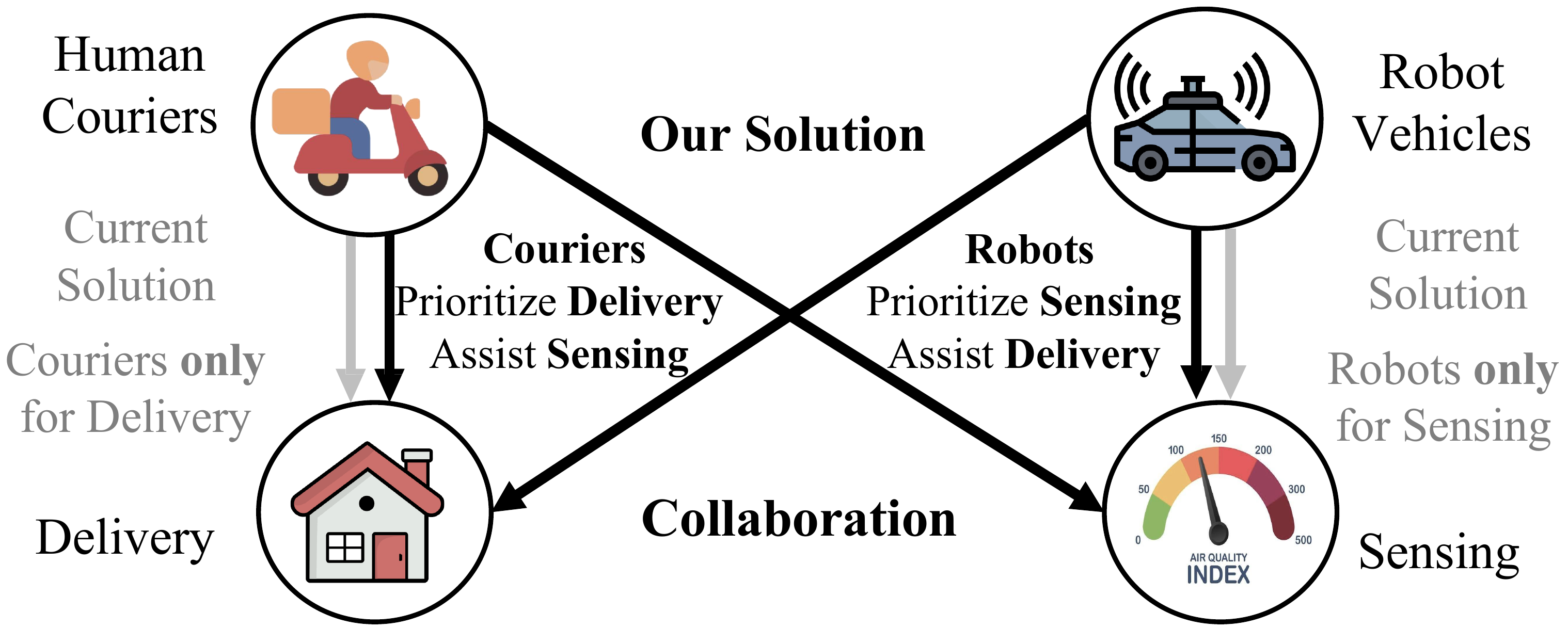}
\caption{Intuition of Human-Robot Collaboration}
\vspace{-20pt}
\label{fig:Idea}
\end{figure}

Rapid advancements in Artificial Intelligence (AI) and sensing technologies have enabled diverse smart city services, such as large-scale real-time order dispatch in food delivery \cite{UberEats} and robot vehicle (RV) cruising for urban sensing \cite{GoogleStreetView}. While many studies \cite{jiang2025hcride,wang2023fairmove,wang2022spatiotemporal} focus on improving the efficiency of individual services, they often treat these services in isolation, overlooking opportunities to leverage resources across services for higher overall efficiency. For instance, food-delivery couriers could collect air-quality or traffic data along their routes, while sensing RVs could assist with food delivery during peak hours, forming a win-win collaboration, as illustrated in Fig.~\ref{fig:Idea}.

Although a collaborative paradigm holds great potential for improving the efficiency of heterogeneous urban services, designing effective algorithms for joint optimization remains nontrivial due to several fundamental \textbf{challenges}: (i) \textit{Conflicting objectives and asynchronous reward feedback.} Different services often have independent and even conflicting objectives, making resource sharing difficult without compromising the performance of individual tasks. Moreover, it is challenging to assess how a single action impacts multiple objectives when rewards are observed asynchronously. For example, the potential sensing benefit of an order dispatch decision cannot be determined in advance, as it depends on subsequent sensing actions. (ii) \textit{Real-time coordination in dynamic environments.} Coordinating large numbers of agents in real time is inherently complex and computationally demanding, particularly in urban systems involving multiple interdependent services. (iii) \textit{Human-robot collaboration.} The coexistence of human workers (e.g., couriers) and autonomous robots (e.g., sensing RVs) adds complexity. Robots follow instructions strictly, whereas human workers have diverse preferences and behaviors. Their satisfaction and willingness to participate should be considered in decision-making to ensure sustainability. 

To address these challenges, we propose \textbf{\M}, a two-layer human-robot collaboration framework for jointly optimizing real-time crowdsourced delivery and urban sensing services. Our objectives are threefold: minimizing overdue orders, maximizing delivery income, and enhancing urban sensing coverage. 
The framework introduces \textbf{hybrid reward-value feedback} to address the first challenge, i.e., asynchronous reward signals in multi-objective optimization. Specifically, the order dispatch layer makes decisions based on both immediate delivery rewards and estimated sensing values computed by the sensing layer, representing expected future sensing returns. To address the second and third challenges, we further design two coupled components in \M:
\textbf{(i)} an efficient \underline{M}ap\underline{R}educe-based \underline{K}-\underline{Sub}modular maximization module for order dispatch, referred to as \textbf{\UM}. This module leverages the submodular property of the matching function and employs the MapReduce paradigm to enable parallel, real-time coordination in large-scale dispatch problems, thereby enhancing computational efficiency. \textbf{(ii)} a \underline{D}eep \underline{S}ubmodular \underline{R}eward \underline{Q}-\underline{N}etwork (\textbf{\DM}) for sensing route planning. Here, human couriers follow their preferred routes to encourage participation, while RVs are guided by a coverage-aware spatio-temporal reward function to dynamically balance sensing and delivery tasks. To facilitate human-robot coordination, sensing gains from human agents are computed directly, whereas those from RVs are passed to the upper layer and aggregated using a submodular function.

To summarize, our main contributions are as follows:
\begin{itemize}
    \item \textbf{Conceptually}, we study the optimization of heterogeneous urban services through human-robot collaboration. In particular, we integrate two originally independent services, food delivery performed by human couriers and urban sensing conducted by RVs, for joint optimization.
    Our motivation is to enable each service to \textbf{\textit{opportunistically}} utilize idle resources to support the other through cooperation.

    \item \textbf{Technically}, we propose \textbf{\M}, a two-layer human-robot collaboration framework designed to jointly optimize heterogeneous urban services. The upper decision layer is a scalable MapReduce-based K-Submodular maximization module, \textit{\textbf{\UM}}, for efficient order dispatch. 
    The lower decision layer includes a novel deep reinforcement learning algorithm called \textit{\textbf{\DM}}, for effective sensing route planning. 

    \item \textbf{Experimentally}, we evaluate \M on a real-world dataset from a food delivery platform, containing 160K orders collected in Shanghai. Experimental results demonstrate that our \M improves sensing coverage by 29.7\% and courier income by 39.2\% on average in most settings, while also significantly reducing the number of overdue orders.
\end{itemize}

\section{Problem Formulation}
\label{sec:Formulation}

Considering the sequential nature of crowdsourced delivery and urban sensing tasks, we formulate their joint optimization as a Markov Decision Process (MDP), denoted by
$\mathcal{M} = (\mathcal{S}, \mathcal{A}, \mathcal{R}, \mathcal{P}, \gamma)$,
where $\mathcal{S}$ is the state space, $\mathcal{A}$ is the action space, $\mathcal{R}$ is the reward function, $\mathcal{P}(s_{t+1}\mid s_t,a_t)$ specifies the state transition dynamics, and $\gamma \in [0,1)$ is the discount factor. In our setting, an entire day is divided into multiple time slots. At each slot, agents interact with the environment following a policy $\pi(a_t|s_t)$ that selects the joint action based on the current state.

\begin{itemize}
    \item \textbf{Agent}: We model human couriers and sensing RVs as agents. When not distinguishing the two types, we use $d^i$ to denote the $i$-th agent. Specifically, the $i$-th courier is $c^i$ and the $i$-th RV is $rv^i$. The total fleet sizes of couriers and RVs are $N^c$ and $N^{rv}$, respectively, and are assumed constant throughout the day. In contrast, the numbers of available couriers and RVs vary over time and are denoted at time step $t$ by $n_t^{c}$ and $n_t^{rv}$.

    \item \textbf{State $\mathcal{S}$}: Each agent has a state $s_t^i \in \mathcal{S}$, defined as the concatenation of three components: agent-specific features $as_t^i$, order-related features $os_t^i$, and sensing-related features $ss_t^i$, i.e., $s_t^i = [as_t^i,\, os_t^i,\, ss_t^i]$. The agent-specific component is $as_t^i = [t,\, g_t^i,\, avail_t^i,\, type^i]$, where $t$ is the current time step, $g_t^i$ is the region (grid) ID where agent $i$ is located, $avail_t^i$ indicates whether the agent is available, and $type^i$ specifies whether the agent is a human courier or an RV. The order-related component $os_t^i = [o_{p,t}^i,\, o_{d,t}^i,\, o_{ddl,t}^i,\, o_{fee,t}^i]$ includes the pickup and drop-off regions, the promised delivery deadline, and the delivery fee. Here, $os_t^i$ corresponds to the order(s) currently assigned to agent $i$ (denoted by $\mathcal{O}_t^i$), with a capacity constraint $|\mathcal{O}_t^i|\le c_d$. The sensing-related component is $ss_t^i = [g_{t}^{i,\mathrm{last}},\, neigh_t^i,\, prop(g_t^i)]$, where $g_{t}^{i,\mathrm{last}}$ denotes the last region where agent $i$ performed sensing, $neigh_t^i$ is the number of adjacent regions visited by agent $i$ in the past hour, and $prop(g_t^i)$ encodes region properties such as region type and accessibility.

    \item \textbf{Action $\mathcal{A}$}: We consider a joint action $a_t=(a_t^o, a_t^s)\in\mathcal{A}$ in our two-layer decision framework, where $\mathcal{A}=\mathcal{A}^o\times\mathcal{A}^s$. In the dispatch layer, $a_t^o$ assigns each available order $o_t^j\in O_t$ to an agent $d^i$ (a courier or an RV). In the sensing layer, we make routing decisions only for RVs: $a_t^s=\{a_{t}^{s,i}\}_{i=1}^{n_t^{rv}}$, where each $a_{t}^{s,i}$ moves an $rv^i$ from its current region to one of the eight neighboring regions (8-connected directions). Couriers do not take routing actions in the sensing layer and are assumed to travel according to their own preferences (e.g., fast and high-reward routes), whereas RVs strictly follow $a_t^s$.

    \item \textbf{Reward $\mathcal{R}$}: 
    At the end of time slot $t$, we define the reward received by each agent $d^i$ during this slot $t$ as $r_t^i = (r_{d,t}^i,\, r_{s,t}^i)$, where $r_{d,t}^i$ and $r_{s,t}^i$ denote the delivery and sensing rewards, respectively. Here, $r_{d,t}^i$ aggregates the delivery rewards from all orders completed by agent $d^i$ during time slot $t$, and the definition is shown as:
    \begin{equation} \label{eq:1}
    r_{d,t}^i=\textstyle\sum_{o^j\in \mathcal{O}_{t}^{i,\mathrm{comp}}} o_{fee,t}^j \times \beta^{\Delta t_{o^j}},
    \end{equation}
    where $\mathcal{O}_{t}^{i,\mathrm{comp}}$ is the set of orders completed by agent $d^i$ during time slot $t$, $o_{fee,t}^j$ is the delivery fee of order $o^j$, $\Delta t_{o^j}$ is its delivery delay, and $\beta \in (0,1)$ is a late-delivery discount factor. If $\Delta t_{o^j}=0$, the full fee is obtained; otherwise, it decays exponentially with $\Delta t_{o^j}$. The sensing reward $r_{s,t}^i$ is defined as:
    \begin{equation} \label{eq:2}
    r_{s,t}^i = r^{\mathrm{reg}}_{g_t^i,d^i} + r^{\mathrm{nbr}}_{g_t^i,d^i} - r^{\mathrm{pen}}_{g_t^i,d^i}.
    \end{equation}
    Here, $g_t^i$ is the region visited by agent $d^i$ at time slot $t$. The regional term $r^{\mathrm{reg}}_{g_t^i,d^i}$ measures the sensing value at $g_t^i$ and is set to zero if $g_t^i$ has been visited within the past hour to avoid redundancy. The neighboring term $r^{\mathrm{nbr}}_{g_t^i,d^i}$ encourages broader coverage by rewarding visits to regions adjacent to unsensed regions, computed in the same way as $r^{\mathrm{reg}}_{g_t^i,d^i}$. The penalty term $r^{\mathrm{pen}}_{g_t^i,d^i}$ imposes a large cost when an order assigned to $d^i$ misses its deadline, discouraging risky sensing actions that may delay deliveries.

    \item \textbf{Transition $\mathcal{P}$}: A mapping $\mathcal{S}\times\mathcal{A}\times\mathcal{S}\rightarrow[0,1]$, where $\mathcal{P}(s_{t+1}\mid s_t,a_t)$ denotes the probability of transitioning from $s_t$ to $s_{t+1}$ under the action $a_t$. In \M, $\mathcal{P}$ is driven by within-step updates of agent locations and order statuses in the Environment and Simulator modules: routing actions $a_t^s$ move RVs across regions, while the dispatch action $a_t^o$ assigns newly arrived orders to both couriers and RVs.

\begin{figure*}[htbp]
    \centering
    \includegraphics[width=0.95\linewidth]{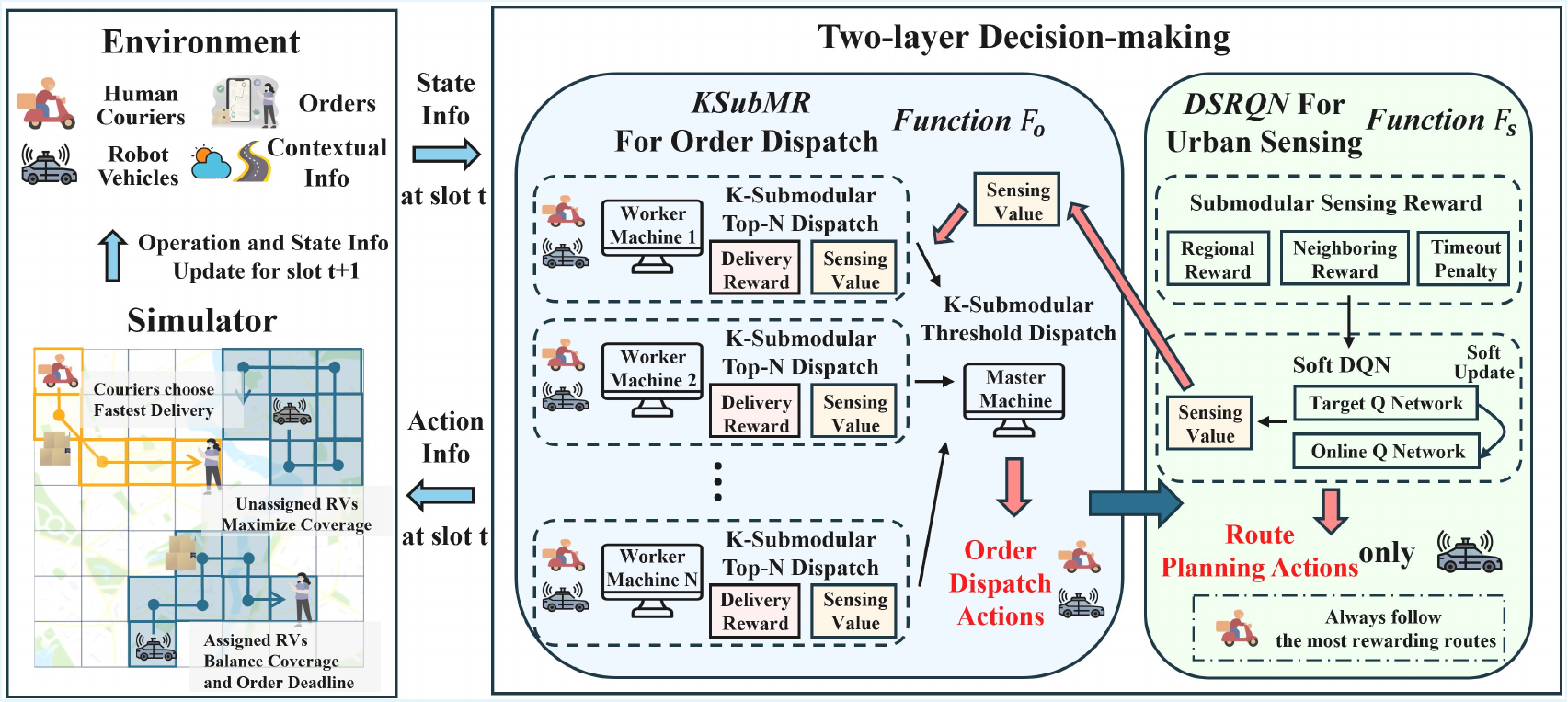}
    \caption{An Overview of the Proposed \M Framework}
    \label{fig:framework}
\end{figure*}

    \item \textbf{Policy $\mathcal{\pi}$}:
    In an MDP, a policy $\pi$ maps states to actions. \M employs two policies in a two-layer decision framework and updates them at each time step: the order dispatch policy $F_o$, which generates the dispatch action $a_t^o$ for all agents, and the urban sensing policy $F_s$, which produces routing actions $a_{t}^{s}$ for RVs.
\end{itemize}

\section{Design of \M}
\label{sec:model_design}
To solve the above MDP, we propose a two-layer human-robot collaboration framework, \textbf{\M}. As shown in Fig.~\ref{fig:framework}, \M comprises the \textit{Environment}, the \textit{Simulator}, and a \textbf{\textit{Decision-making}} module with two coupled layers, order dispatch and route planning. In the \textbf{order dispatch} layer, the system assigns orders to both human couriers and RVs. In the \textbf{route planning} layer, couriers follow their preferences (i.e., profit-maximizing routes), whereas RVs strictly follow system-generated actions for sensing.

\subsection{Environment and Simulator} \label{sec:Environment and Simulator}

In the \textit{Environment} module, we model couriers, RVs, and orders, together with contextual factors such as weather and the road network. Each order includes pickup and drop-off regions, a delivery fee, and a deadline. Both couriers and RVs can perform delivery and sensing, but with different roles: couriers prioritize timely deliveries and passively collect sensing data along the way, whereas RVs prioritize sensing to maximize coverage in each period (e.g., hourly). When courier capacity is insufficient, RVs assist with deliveries to reduce overdue orders. We also incorporate weather and road-network conditions, which affect agent behavior, travel speed, and region accessibility.

The \textit{Simulator} models agent operations from time slot $t$ to $t{+}1$ on a grid-based map. As shown in Fig.~\ref{fig:framework}, we capture three behavior patterns: (1) Couriers always take reward-maximizing routes (modeled as the fastest routes), reflecting their preference for minimum travel time, and sense passively; (2) RVs without delivery tasks follow system-generated sensing actions to maximize coverage; and (3) RVs with delivery tasks follow sensing actions under deadlines, balancing coverage and timely delivery.

\subsection{Efficient Two-layer Decision-making}

In this section, we present the details of the proposed efficient two-layer decision-making strategy with human-robot collaboration, which comprises two coupled layers: (i) the \textbf{upper layer}, an efficient MapReduce-based K-submodular maximization module for order dispatch (\UM), and (ii) the \textbf{lower layer}, a Deep Submodular Reward Q-Network (\DM) for route planning. Both couriers and RVs accept order dispatch actions, but only RVs strictly follow system-generated route planning actions.

\subsubsection{\underline{Optimization Objectives}}\label{sec:Optimization_Objectives}

\begin{figure*}[htbp]
    \centering
    \includegraphics[width=0.95\textwidth]{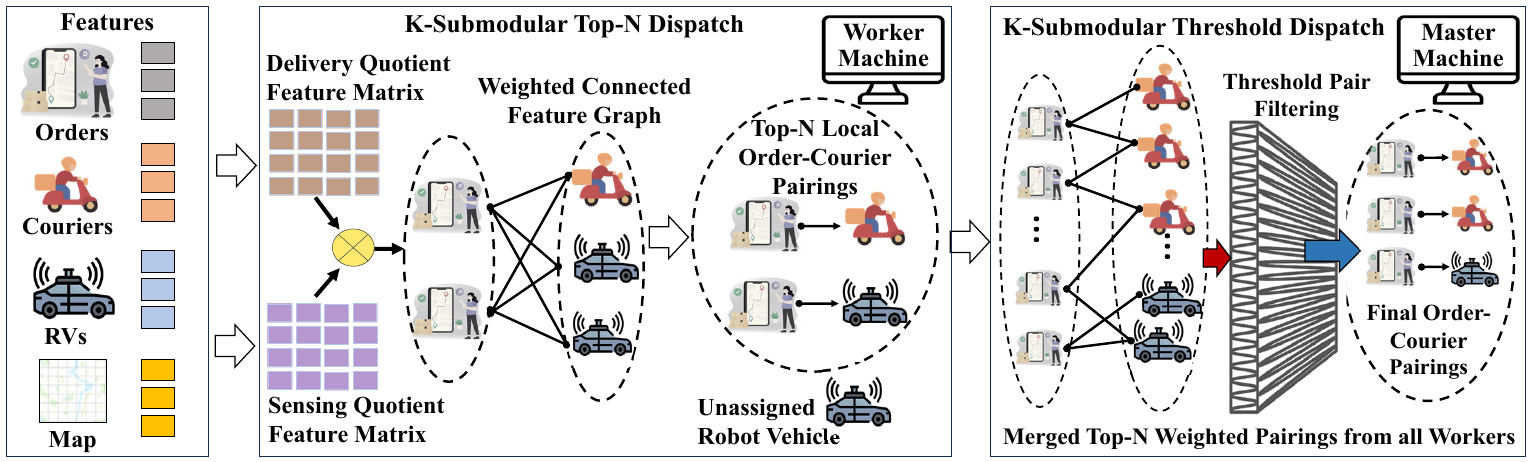}
    \caption{An Overview of the Proposed \UM}
    \label{fig:workflow}
    \vspace{-10pt}
\end{figure*}

At the beginning of each time slot $t$, there are $n_t^{c}$ available human couriers and $n_t^{rv}$ available RVs. Let $D_t={d^i}_{i=1}^{n_t^{c}+n_t^{rv}}$ denote the set of available agents and $O_t$ denote the set of pending orders at time $t$.

In our two-layer human--robot collaboration framework, the upper layer uses the order dispatch policy $F_o$ (implemented as \UM) to assign orders to agents by maximizing delivery and sensing rewards, while the lower layer uses the route-planning policy $F_s$ (implemented as \DM) to maximize RV sensing rewards subject to on-time delivery. Moreover, $F_o$ relies on the sensing value estimated by $F_s$. As a result, $F_o$ outputs the dispatch action $a_t^o$ (i.e., the matching set $S_t$) and $F_s$ outputs the routing actions $a_t^s$ for RVs, forming the joint action $a_t=(a_t^o,a_t^s)$ in our MDP.

The upper-layer function $F_o$ is defined as a binary assignment indicator:
\begin{equation} \label{eq:3}
F_o: O_t \times D_t \rightarrow \{0,1\},
\end{equation}
where $F_o(o, d) = 1$ indicates that order $o \in O_t$ is assigned to agent $d \in D_t$. The resulting assignment set is
\begin{equation} \label{eq:4}
S_t = \{(o, d) \in O_t \times D_t \mid F_o(o, d) = 1\}.
\end{equation}
we score each candidate pair $(o,d)$ by
\begin{equation} \label{eq:5}
r(o, d) = r_d(o,d) + v_s(o,d),
\end{equation}
where $o \in O_t$ and $d \in D_t$. Here, $r_d(o,d)$ is the delivery reward for completing order $o$ (computed following Eq.~\ref{eq:1}), and $v_s(o,d)$ is the urban sensing value calculated in Section~\ref{sec:DSRQN for Urban Sensing}. The resulting reward-driven many-to-many matching is formulated as the following optimization problem:
\begin{equation} \label{eq:6}
\begin{aligned}
S_t \gets & \: \arg\max_{S_t \subseteq O_t \times D_t} \quad  R_d(S_t) + V_s(S_t) \\
 \text{s.t.} \quad 
& \textstyle\sum_{d \in D_t} \mathbf{1}_{(o, d) \in S_t} \leq 1, \quad \forall o \in O_t, \\
& \textstyle\sum_{o \in O_t} \mathbf{1}_{(o, d) \in S_t} \leq c_d, \quad \forall d \in D_t,
\end{aligned}
\end{equation}
where
\begin{align}
R_d(S_t) &= \textstyle\sum_{(o,d)\in S_t} r_d(o, d), \nonumber\\
V_s(S_t) &= f_{\text{sub}} \Big(\textstyle\sum_{(o,d)\in S_t} v_s(o, d)\Big). \label{eq:7}
\end{align}

Here, $R_d(S_t)$ is the delivery reward of the assignment set $S_t$, $V_s(S_t)$ is its overall sensing value, and $f_{\text{sub}}(\cdot)$ is a submodular aggregation function. A key challenge in optimizing $F_o$ in Eq.~\ref{eq:3} is \textbf{asynchronous feedback}: $R_d(S_t)$ is available once $S_t$ is decided, whereas $V_s(S_t)$ cannot be observed at dispatch time because it depends on subsequent RV routing actions. Thus, the objective in Eq.~\ref{eq:6} is coupled with future decisions, making $F_o$ nontrivial to optimize.

To address this issue, we propose \M with \textbf{hybrid reward--value feedback}. Specifically, the upper layer \UM applies the order dispatch policy $F_o$ to solve Eq.~\ref{eq:6} by combining the immediate reward $R_d(S_t)$ with the estimated value $V_s(S_t)$. The sensing value terms $v_s(o,d)$ and $V_s(S_t)$ are computed in the lower layer by \DM, and are fed back to \UM for dispatch optimization.

\subsubsection{\underline{\UM for Order Dispatch Decision-making}}\label{sec:MRKS-Dispatch for Order Dispatch}
The multi-objective optimization in Eq.~\ref{eq:6} can be formulated as a weighted bipartite graph for many-to-many matching, for which the Kuhn--Munkres algorithm~\cite{guo2021concurrent} is a common solution. However, as the numbers of orders, couriers, and RVs grow substantially, its \textit{computational complexity} increases sharply. Coupled with the computational cost of deep RL, this approach becomes impractical for meeting the real-time requirements of order dispatch systems.

To address the computational-efficiency challenge and enable real-time coordination in highly dynamic environments, we propose the \UM algorithm, which integrates a two-round MapReduce framework with $k$-submodular maximization, as illustrated in Fig.~\ref{fig:workflow}. This approach leverages the submodular property~\cite{nemhauser1978analysis} of the dispatch policy function $F_o$, which provides provable approximation guarantees in combinatorial optimization~\cite{mirzasoleiman2016distributed}. Moreover, the composability of local solutions makes it well-suited for parallelization in distributed frameworks such as MapReduce~\cite{dean2008mapreduce}.

In the first MapReduce round, each \textit{worker machine} runs the K-Submodular Top-$N$ Dispatch algorithm (Algorithm~\ref{alg:topN}) on its partition of orders and agent states $\{s_t^i\}$ to produce a local assignment set $S_j$. Here, $M$ denotes the number of parallel workers. In the second MapReduce round, the \textit{master machine} aggregates the worker outputs, reconstructs a bipartite graph over feasible order--agent pairs, and applies the K-Submodular Threshold Dispatch algorithm (Algorithm~\ref{alg:thd}) to generate the final dispatch.

\textbf{K-Submodular Top-$N$ Dispatch (Algorithm~\ref{alg:topN})}:
During the first MapReduce round, orders are distributed to worker machines with agent states $s_t^i$. Two quotient matrices are constructed, namely the Delivery Quotient ($DQ$) and Sensing Quotient ($SQ$), based on $r_d(o,d)$ and $v_s(o,d)$. Guided by these matrices, each order iteratively selects its top-$N$ agents in a round-robin manner. Agents are ranked by the computable dispatch reward $R_d(\cdot)$ and the sensing value $V_s(\cdot)$ from \DM, where $\Delta_{o,d}(S_j)$ in Algorithm~\ref{alg:topN} denotes the marginal gain of adding $(o,d)$ to the local assignment set $S_j$ under $R_d(\cdot)+V_s(\cdot)$. The $k$-submodular structure reduces redundancy in local solutions and yields disjoint assignments. Each worker then transmits its results to the master node.

\begin{algorithm}[!t]
\caption{K-Submodular Top-$N$ Dispatch}
\label{alg:topN}
\begin{algorithmic}[1]
\State \textbf{Input:} dispatch reward function $R_d(\cdot)$, sensing value function $V_s(\cdot)$, agent limit $N$ per order
\State \textbf{Output:} a sparse pair set $S \subseteq O \times D$
\State $S \gets \emptyset$
\For{each $o \in O$}
    \State $S_j \gets \emptyset$
    \For{$i = 1$ to $N$}
        \State $(o,d) \gets \arg\max_{d \in D} \Delta_{o,d}(S_j)$
        \State $S_j \gets S_j \cup \{(o,d)\}$
    \EndFor
    \State $S \gets S \cup S_j$
\EndFor
\State \Return $S$
\end{algorithmic}
\end{algorithm}

\begin{algorithm}[!t]
\caption{K-Submodular Threshold Dispatch}
\label{alg:thd}
\begin{algorithmic}[1]
\State \textbf{Input:} value oracle for monotone $k$-submodular $f_{\text{sub}}$, $O$, $D$, $\{c_d\}_{d\in D}$, $\epsilon\in(0,1)$
\State \textbf{Output:} $S \subseteq O \times D$
\State $S \gets \emptyset,\ \{S_d\}_{d\in D}\gets \emptyset$
\State $\tau \gets \tau_0 \gets \max \{ r_d(o,d) : o \in O, d \in D \}$
\While{$\tau > \epsilon \tau_0$}
  \ForAll{$(o,d)\in O\times D$}
    \If{$(o,d)\notin S$ \textbf{and} $f_{\text{sub}}(S \cup \{(o,d)\}) - f_{\text{sub}}(S) \ge \tau$ \textbf{and} $|S_d| < c_d$}
      \State $S_d \gets S_d \cup \{(o,d)\},\ S \gets S \cup \{(o,d)\}$
    \EndIf
  \EndFor
  \State $\tau \gets (1-\epsilon)\tau$
\EndWhile
\State \Return $S$
\end{algorithmic}
\end{algorithm}

\textbf{K-Submodular Threshold Dispatch (Algorithm~\ref{alg:thd})}:
In the second MapReduce round, the master machine aggregates the worker outputs and reconstructs a bipartite graph connecting orders and agents to regenerate the quotient matrices. The final dispatch is obtained by applying a decreasing sequence of thresholds, starting from a selected order--agent pair. For each threshold $\tau$, the algorithm scans unassigned order--agent pairs and selects those whose marginal gains exceed $\tau$. This process continues until all orders are assigned or the threshold falls below a predefined value.

\subsubsection{\underline{\DM for Urban Sensing Decision-making}}\label{sec:DSRQN for Urban Sensing}

We now present our Deep Submodular Reward Q-Network (\DM), which serves two objectives: (1) generating RV route-planning actions $a_t^s$ to improve sensing coverage, and (2) estimating the aggregated sensing value $V_s(S_t)$ for the order dispatch layer. Specifically, \DM builds on a Q-value-based RL algorithm to estimate the expected return of routing actions and guide RV behavior.

\textbf{Firstly, we describe how route-planning actions are generated.} We introduce key concepts in \DM. The Q-value for a state--action pair, denoted as $Q(s_t^i, a_{t}^{s,i})$, represents the expected cumulative sensing reward when RV $rv^i$ takes routing action $a_{t}^{s,i}$ at state $s_t^i$. Accordingly, we define
\begin{equation}
    Q(s_t^i, a_{t}^{s,i})
    = \mathbb{E}\!\left[\sum_{k=0}^{H} r_{s,t+k}^i \,\bigg|\, s_t^i, a_{t}^{s,i}\right].
\end{equation}

In \DM, we parameterize the Q-function as $Q(s_t^i, a_{t}^{s,i}; \theta)$, where $\theta$ denotes the parameters of $F_s$. The sensing reward $r_{s,t}^i$ in Eq.~\ref{eq:2} serves as the learning signal for estimating $Q(s_t^i, a_{t}^{s,i})$ and consists of three components: the regional reward $r^{\mathrm{reg}}_{g_t^i,d^i}$, the neighboring reward $r^{\mathrm{nbr}}_{g_t^i,d^i}$, and the timeout penalty $r^{\mathrm{pen}}_{g_t^i,d^i}$.

The optimal Q-value $Q^*(s_t^i, a_{t}^{s,i})$ represents the maximum expected return achievable from state $s_t^i$ by taking action $a_{t}^{s,i}$ and then making optimal routing decisions thereafter. In \DM, we learn a parameterized Q-function $Q(s_t^i, a_{t}^{s,i}; \theta)$ and derive the route-planning policy function $F_s$ via a greedy action-selection rule:
\begin{equation} \label{eq:9}
    F_s(s_t^i;\theta)
    = \arg\max_{a_{t}^{s,i}}\, Q(s_t^i, a_{t}^{s,i}; \theta).
\end{equation}
Finally, $F_s$ outputs the route-planning action $a_{t}^{s,i}$ for each RV $rv^i$ in the lower decision layer.

\textbf{Next, we describe the computation of the aggregated sensing value $V_s(S_t)$}, which is passed to \UM for solving $F_o$. We first define the sensing value $v_s(o,d)$ for an order--agent pair $(o,d)\in S_t$, which estimates the sensing gain of assigning order $o$ to agent $d$ (an RV in our setting, since only RVs follow system-generated routing actions). For illustration, we use an order $o^j$ and an agent $d^i$ as a running example. We adopt an RL value-function view to quantify the expected future sensing return from the agent's current state under the learned routing policy.

Since sensing is location-dependent, we define the region-level sensing value $V_{\mathrm{reg}}$ at region $g$ given agent state $s_t^i$ as
\begin{equation}
V_{\mathrm{reg}}(g \mid s_t^i)
= Q\!\left(s_t^i,\, F_s(s_t^i;\theta);\theta\right),
\end{equation}
where $F_s$ is the route-planning function in Eq.~\ref{eq:9}. Given an order $o_t^j \in O_t$ and an agent $d^i \in D_t$, let $g_t^i$ be the current region of $d^i$ and $o_{d,t}^j$ the drop-off region of $o_t^j$. We estimate the sensing value of assigning $o_t^j$ to $d^i$ as
\begin{equation}
v_s(o_t^j,d^i)
= V_{\mathrm{reg}}\!\left(o_{d,t}^j \mid s_t^{i}\right)
- V_{\mathrm{reg}}\!\left(g_t^{i} \mid s_t^{i}\right).
\end{equation}
This value is computed as the difference between the region-level values at the drop-off region and the current region.

After obtaining $v_s(o_t^j,d^i)$ for each selected pair $(o_t^j,d^i)\in S_t$, we aggregate them to compute $V_s(S_t)$ in Eq.~\ref{eq:7}. To capture diminishing returns caused by spatial redundancy, we adopt a submodular aggregation function $f_{sub}(\cdot)$ and define
\begin{equation} \label{eq:13} \small
V_s(S_t) \triangleq f_{sub}(S_t)
= \left(\sum_{(o_t^j,d^i)\in S_t} v_s(o_t^j,d^i)\right)\!\cdot\! \left(1-\rho(S_t)\right),
\end{equation}
where the redundancy penalty $\rho(S_t)$ is
\begin{equation} \label{eq:14}\small
\rho(S_t) = \frac{1}{|S_t|(|S_t|-1)}
\sum_{\substack{m \ne n}}
\exp\!\left( -\frac{\mathrm{dis}\big((o^m,d^m),(o^n,d^n)\big)}{2\sigma^2} \right).
\end{equation}

Here, $\rho(\cdot)$ measures spatial redundancy among the paths induced by the assignments in $S_t$~\cite{lin2011class}. The term $\mathrm{dis}\big((o^m,d^m),(o^n,d^n)\big)$ denotes the distance between the $m$-th and $n$-th order--agent assignments, and the exponential kernel assigns higher similarity to closer (more redundant) paths. The normalization by $|S_t|(|S_t|-1)$ keeps $\rho(S_t)$ scale-invariant, while the bandwidth $\sigma$ controls the effective influence radius. The submodularity of $f_{sub}(\cdot)$ originates from the normalized pairwise similarity measure $\rho(\cdot)$, which induces diminishing returns in the aggregation in Eq.~\ref{eq:13}.

\section{Evaluation}

In this section, we conduct extensive experiments to evaluate the performance of our \M.

\subsection{Evaluation Setup}

\subsubsection{Parameter Setting and Experiment Environment}
We evaluate \M under fine-grained \textbf{spatial} and \textbf{temporal} settings to improve practicality. Specifically, we set the region size (GS) to $(0.05\,\mathrm{km})^2$ and use a 1-minute time slot to support timely delivery and high-resolution sensing. We run experiments on 40 Intel(R) Xeon(R) Gold 5218R CPU @ 2.10GHz processors. We implement \DM in TensorFlow~2.4.1 and Python~3.8, and train on an NVIDIA TITAN RTX (TU102) GPU using real-world data distributions. Training uses an experience replay buffer of size $10^6$ and a batch size of 128. We optimize Q-networks with Adam (learning rate $0.01$) and set the discount factor $\gamma$ to $0.9$.

\begin{figure*}[t]
  \centering
  \begin{subfigure}[t]{0.22\textwidth}
    \centering
    \includegraphics[width=\linewidth]{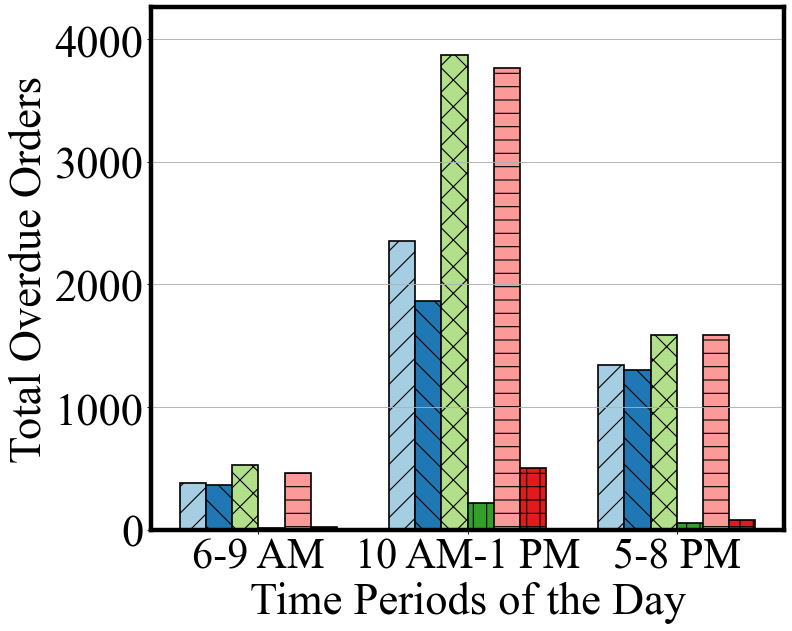}
    \caption{500 RVs}
    \label{fig:overdue_500}
  \end{subfigure}
  \hfill
  \begin{subfigure}[t]{0.22\textwidth}
    \centering
    \includegraphics[width=\linewidth]{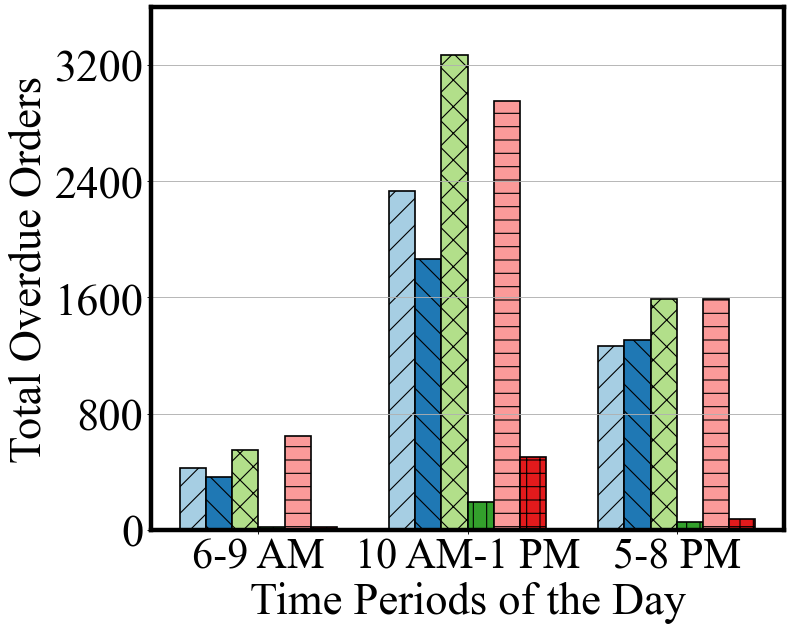}
    \caption{1000 RVs}
    \label{fig:overdue_1000}
  \end{subfigure}
  \hfill
  \begin{subfigure}[t]{0.22\textwidth}
    \centering
    \includegraphics[width=\linewidth]{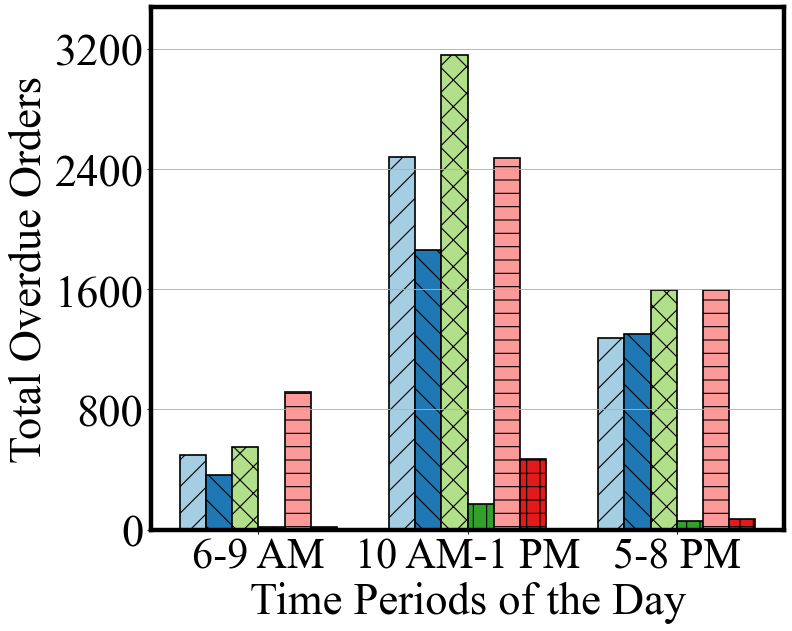}
    \caption{2000 RVs}
    \label{fig:overdue_2000}
  \end{subfigure}
  \hfill
  \begin{subfigure}[t]{0.22\textwidth}
    \centering
    \includegraphics[width=\linewidth]{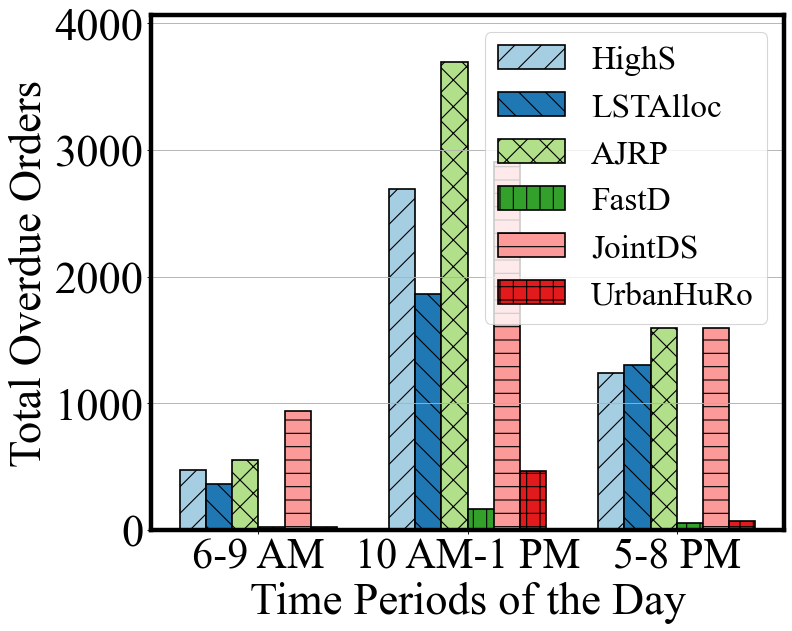}
    \caption{4000 RVs}
    \label{fig:overdue_4000}
  \end{subfigure}
  \caption{Overdue orders across three key time periods, i.e., morning rush hours, noon peak hours, and evening rush hours.}
\label{fig:exp_results_overdue}
\end{figure*}

\subsubsection{Evaluation Metrics}
We evaluate the joint optimization of order dispatch and urban sensing using three metrics: number of overdue orders, sensing coverage, and the average courier income. These metrics are defined as follows:

\begin{itemize}
    \item \textbf{Overdue Orders}: The number of orders not delivered by their deadlines during an hour. 
    
    \item \textbf{Normalized Sensing Coverage ($\mathcal{C}_{\mathcal{N}}$)}: The ratio of the number of visited locations achieved by a model to that achieved by \M.
    
    \item \textbf{Average Courier Income}: Average hourly income of human couriers, calculated based on total revenue from assigned orders minus overdue penalties. 
\end{itemize}

\subsubsection{Datasets} \label{sec:data} We utilize a real-world dataset from a food delivery platform for evaluation~\cite{jiang2023faircod}. The dataset includes over 160K on-demand food delivery orders collected within a week in Shanghai, with around 2,200 active couriers at noon. It contains detailed information such as user IDs, pickup/drop-off locations with timestamps, delivery deadlines, and fees.

\begin{figure*}[t]
  \centering
  \begin{subfigure}[t]{0.22\textwidth}
    \centering
    \includegraphics[width=\linewidth]{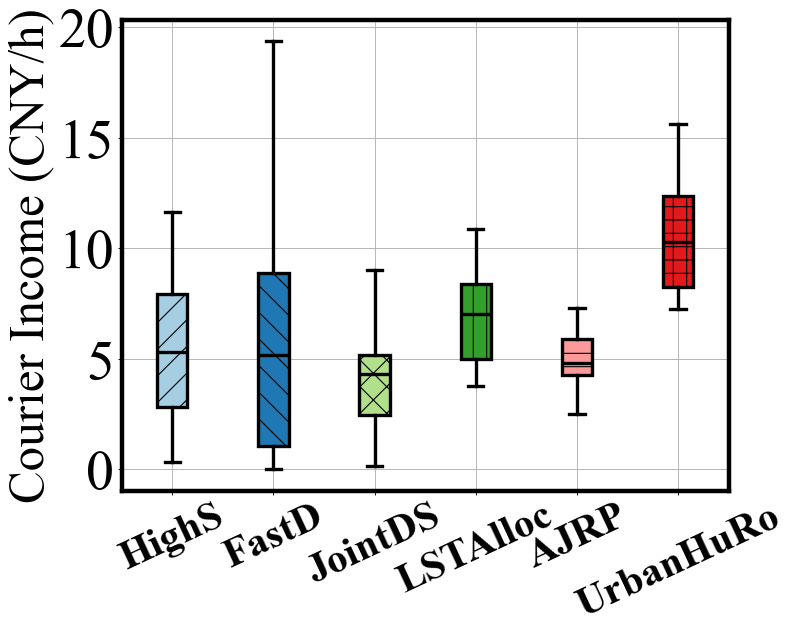}
    \caption{500 RVs}
    \label{fig:incentive_500}
  \end{subfigure}
  \hfill
  \begin{subfigure}[t]{0.22\textwidth}
    \centering
    \includegraphics[width=\linewidth]{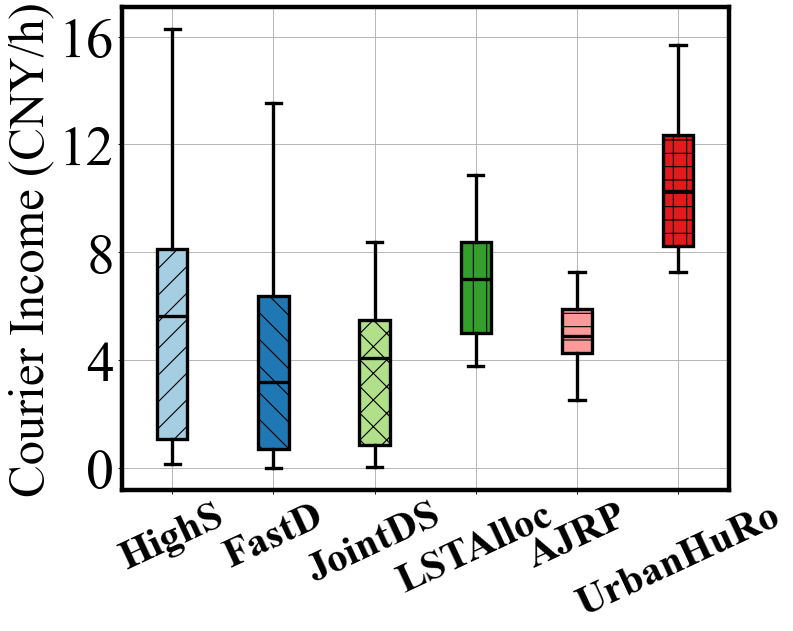}
    \caption{1000 RVs}
    \label{fig:incentive_1000}
  \end{subfigure}
  \hfill
  \begin{subfigure}[t]{0.22\textwidth}
    \centering
    \includegraphics[width=\linewidth]{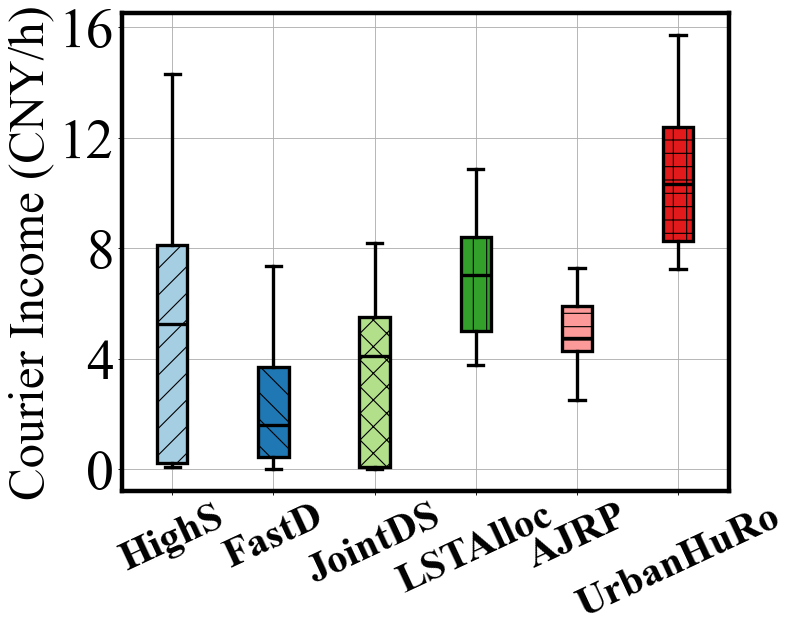}
    \caption{2000 RVs}
    \label{fig:incentive_2000}
  \end{subfigure}
  \hfill
  \begin{subfigure}[t]{0.22\textwidth}
    \centering
    \includegraphics[width=\linewidth]{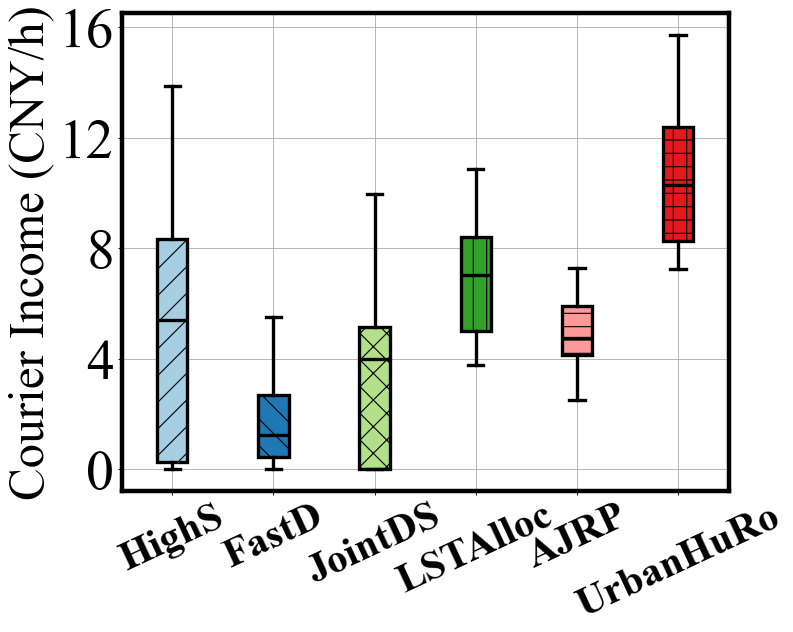}
    \caption{4000 RVs}
    \label{fig:incentive_4000}
  \end{subfigure}
  \caption{Average hourly income of couriers under different numbers of RVs.}
  \vspace{-10pt}
  \label{fig:exp_results_income} 
\end{figure*}

\subsubsection{Baselines}
We compare \M with the following five state-of-the-art baselines.
\begin{itemize}
    \item \textbf{Fastest-Delivery (FastD)}: This approach always selects the nearest agent and performs RV sensing using the Shortest Diversified Path Routing (SDPR) method, based on the K-Shortest Diversified Path Routing algorithm~\cite{lai2022optimized} with $k=1$.

    \item \textbf{High-Sensing (HighS)}: This approach selects the agent that provides the maximum coverage of unvisited locations along its route. It performs sensing using the Trajectory Scheduling for Maximum Task Coverage (TSMTC) algorithm~\cite{fan2019joint}, which selects the neighboring location with the highest coverage potential.

    \item \textbf{Joint-Delivery and Sensing Efficiency (JointDS)}~\cite{xiang2021reusing}: This approach selects the agent with the highest ratio of sensed regions to driving distance. It utilizes the REASSIGN~\cite{lesmana2019balancing} routing method for RVs.

    \item \textbf{LSTAlloc}~\cite{xiang2023lstaloc}: This approach aims to maximize courier profit by prioritizing order assignments based on a balance between current income and driving distance. Any unassigned orders are then allocated to RVs.

    \item \textbf{AJRP}~\cite{liu2023demand}: This approach estimates a cost-to-go function based solely on sensing gain, considering human couriers first. If no couriers are available, it applies the same strategy to assign active orders to RVs.
\end{itemize}

\subsection{Evaluation Results}

\begin{table}[!h]\scriptsize
\caption{Overall Comparison.}
\label{tbl:results}
\centering
\renewcommand{\arraystretch}{1.5}
\setlength{\tabcolsep}{2pt}
\adjustbox{max width=\linewidth}{
\begin{tabular}{c|cccc|cccc|cccc}
\hline
\textbf{\# RVs} & \textbf{500} & \textbf{1000} & \textbf{2000} & \textbf{4000} 
& \textbf{500} & \textbf{1000} & \textbf{2000} & \textbf{4000}
& \textbf{500} & \textbf{1000} & \textbf{2000} & \textbf{4000} \\
\cline{1-13}
\textbf{Models} & \multicolumn{4}{c|}{\textbf{Overdue Orders}} 
& \multicolumn{4}{c|}{\textbf{Coverage ($\mathcal{C}_{\mathcal{N}}$)}} 
& \multicolumn{4}{c}{\textbf{Courier Income}} \\
\hline
\textbf{HighS}       & 291 & 287 & 284 & 281 & 1.01 & 0.79 & 0.72 & 0.81 & 5.76 & 5.55 & 5.07 & 5.00 \\
\textbf{FastD}       & 16  & 15  & 14  & 13  & 0.05 & 0.03 & 0.02 & 0.02 & 5.63 & 4.39 & 3.14 & 2.64 \\
\textbf{JointDS}     & 386 & 374 & 361 & 356 & 0.94 & 0.76 & 0.52 & 0.50 & 4.48 & 4.03 & 3.64 & 3.17 \\
\textbf{LSTAlloc}    & 266 & 266 & 266 & 265 & 0.83 & 0.59 & 0.46 & 0.40 & 8.15 & 8.15 & 8.15 & 8.15 \\
\textbf{AJRP}        & 390 & 374 & 366 & 361 & 0.88 & 0.62 & 0.48 & 0.42 & 5.71 & 5.69 & 5.61 & 5.48 \\
\textbf{\M}     &  31 &  30 &  28 &  28 & 1.00 & 1.00 & 1.00 & 1.00 &11.34 &11.35 &11.34 &11.36 \\
\hline
\end{tabular}
}
\end{table}

We provide an overall comparison of our experimental results in Table~\ref{tbl:results}. As shown, UrbanHuRo consistently achieves strong performance across different RV fleet-size settings in all the three metrics. The details are shown as below.

\subsubsection{Overdue Orders}
Fig.~\ref{fig:exp_results_overdue} compares the number of overdue orders at different times of the day, demonstrating the effectiveness of \M in minimizing delayed deliveries under dynamic, high-demand conditions. \M consistently matches \fd in maintaining timely fulfillment, even during peak periods. In contrast, other baselines exhibit noticeable spikes in overdue orders, particularly around noon, suggesting limited scalability for real-time dispatch.

Quantitatively, \M achieves an orders-of-magnitude reduction in overdue orders compared to state-of-the-art baselines, especially during core business hours (8:00 AM--8:00 PM) when system load is highest. This indicates that \M prioritizes time-sensitive dispatch without sacrificing sensing coverage. Unlike \bs and \jd, which favor coverage at the expense of on-time delivery, \M strikes a better balance, making it more suitable for practical joint delivery and sensing.

\subsubsection{Sensing Coverage} 
As shown in Table~\ref{tbl:results}, with 500 RVs, \bs shows similar map coverage to \M. However, as the number of RVs increases, \M outperforms all baselines. Notably, with 1000, 2000, and 4000 RVs, \M exceeds the best-performing baseline, \bs, by an average of 29.7\% in coverage. This demonstrates \M's scalability and adaptability in effectively balancing urban sensing with delivery efficiency.

\subsubsection{Courier Income} Figure~\ref{fig:exp_results_income} shows that \M consistently boosts earnings for human couriers. On average, \M more than doubles courier income compared to \bs and \jd, and exceeds LSTAlloc, a SOTA method specifically designed to maximize courier earnings, by about 39.2\% in average. This improvement stems from \M's ability to reduce overdue orders, which would otherwise diminish courier compensation. Importantly, \M achieves these income gains while maintaining low overdue orders and strong sensing coverage.

\begin{figure}[!h]
  \centering
  \begin{subfigure}[t]{0.23\textwidth}
    \centering
    \includegraphics[width=\linewidth]{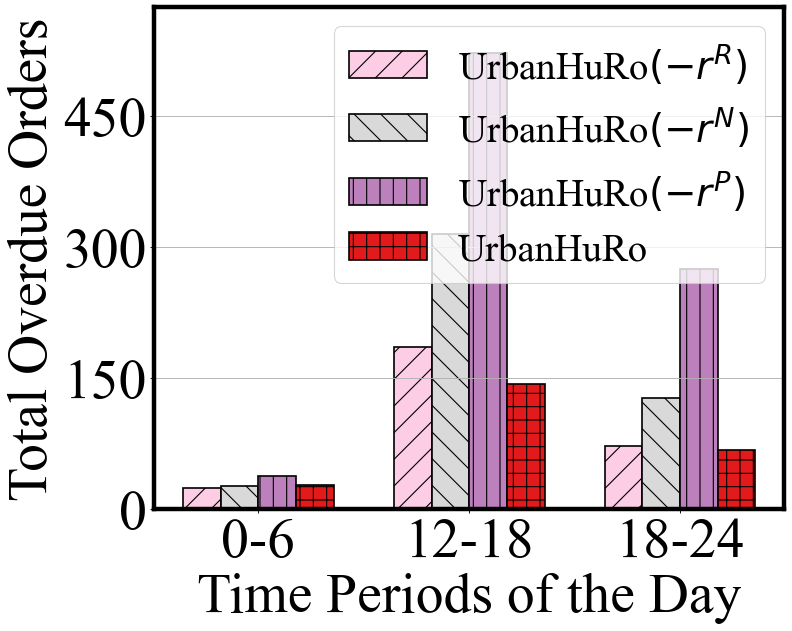}
    \caption{Overdue Orders (1000 RVs)}
    \label{fig:abl_overdue_1000}
  \end{subfigure}
  \hfill
  \begin{subfigure}[t]{0.23\textwidth}
    \centering
    \includegraphics[width=\linewidth]{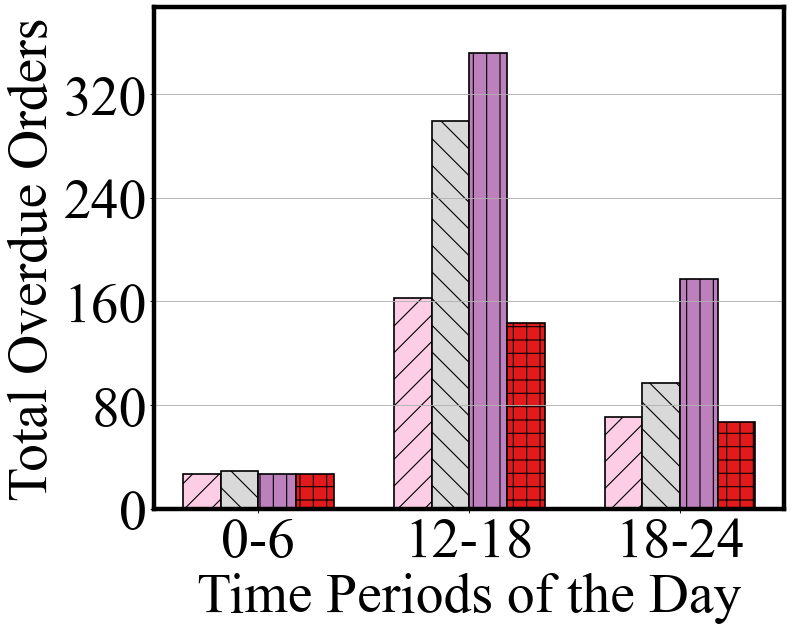}
    \caption{Overdue Orders (2000 RVs)}
    \label{fig:abl_overdue_2000}
  \end{subfigure}
  \hfill
  \begin{subfigure}[t]{0.23\textwidth}
    \centering
    \includegraphics[width=\linewidth]{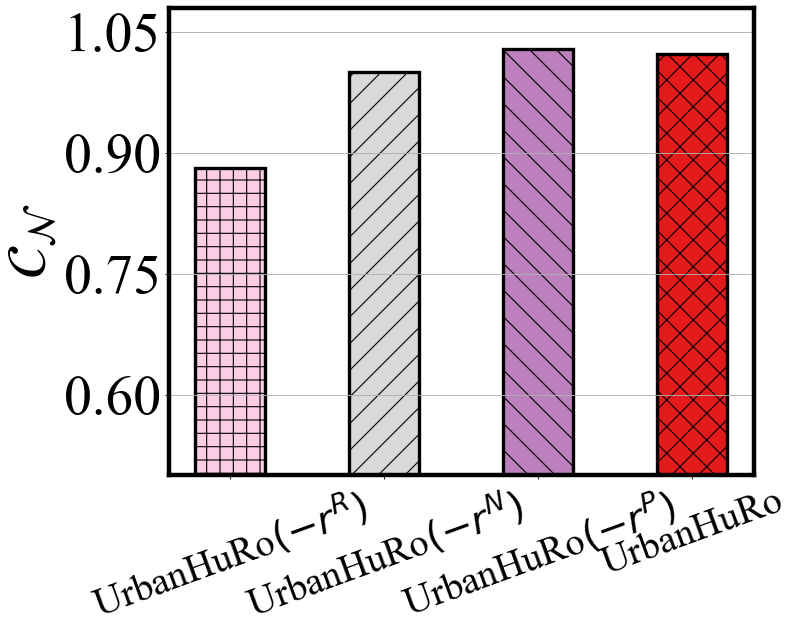}
    \caption{Coverage (1000 RVs)}
    \label{fig:abl_coverage_1000}
  \end{subfigure}
  \hfill
  \begin{subfigure}[t]{0.23\textwidth}
    \centering
    \includegraphics[width=\linewidth]{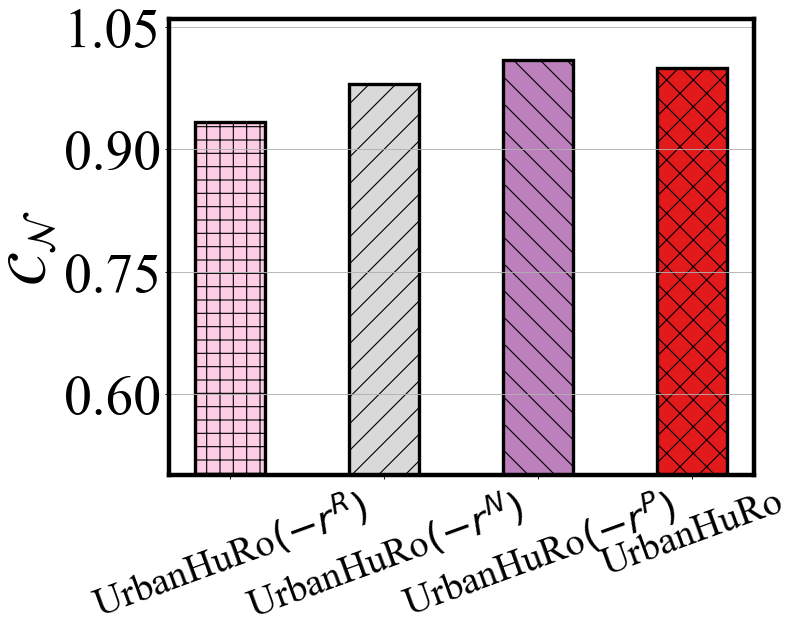}
    \caption{Coverage (2000 RVs)}
    \label{fig:abl_coverage_2000}
  \end{subfigure}
  \caption{Ablation Studies}
  \label{fig:ablation}
  \vspace{-10pt}
\end{figure}

\subsubsection{Ablation Study}
We evaluate three variants of \M: (i) $\M(-r^{\mathcal{R}})$, excluding the regional reward $r^{\mathrm{reg}}_{g^i_t,d^i}$; (ii) $\M(-r^{\mathcal{N}})$, excluding the neighboring reward $r^{\mathrm{nbr}}_{g^i_t,d^i}$; and (iii) $\M(-r^{\mathcal{P}})$, excluding the timeout penalty $r^{\mathrm{pen}}_{g^i_t,d^i}$.
In Figures~\ref{fig:abl_overdue_1000} and~\ref{fig:abl_overdue_2000}, we observe that removing any of the three components leads to an increase in overdue orders. Notably, omitting the timeout penalty $r^{\mathrm{pen}}_{g^i_t,d^i}$ causes the most significant rise in overdue orders throughout the day, whereas removing the neighboring reward $r^{\mathrm{nbr}}_{g^i_t,d^i}$ has minimal impact. In Figures~\ref{fig:abl_coverage_1000} and~\ref{fig:abl_coverage_2000}, we also examine the impact of the three components on sensing coverage. The results show that removing either the regional reward $r^{\mathrm{reg}}_{g^i_t,d^i}$ or the neighboring reward $r^{\mathrm{nbr}}_{g^i_t,d^i}$ weakens the system's sensing capability. In contrast, omitting the timeout penalty $r^{\mathrm{pen}}_{g^i_t,d^i}$ enhances coverage, as the system no longer needs to sacrifice sensing gains to meet delivery deadlines.

\section{Related Work}
\subsection{Urban Service Optimization}

Urban service optimization has attracted much interest from both the research community and industry due to its importance in reducing operational costs, increasing profits, and improving the quality of life. Recent studies have investigated effective order dispatch in ridesharing and logistics \cite{zhang2024nondbrem, jiang2025hcride, jiang2023faircod,9197313,makino2024online}, route planning for urban sensing \cite{ruckin2023informative,ji2025route}, next location recommendation~\cite{xu2025geogen}, and resource allocation and management in mobility systems \cite{wang2023foretaxi, jiang2025uncertainty, tan2024human, he2020data, ruan2020dynamic}. While these works achieve substantial gains within individual services, they often overlook the potential synergies and reciprocal benefits that could be realized through the joint optimization of heterogeneous urban services.

\subsection{Human-Robot Collaboration in Smart Cities}
With the rapid advancement of smart cities, human–robot collaboration has become a key research focus for diverse urban services. For instance, in smart mobility and transportation, prior studies have examined interactions between delivery robots and human drivers in public transit systems \cite{de2024sustainable}. In urban sensing and inspection, robots have been utilized to collaborate with municipal workers and local residents to collect sensing data \cite{munasinghe2024comprehensive}. Researchers have explored human–robot collaboration for public safety and emergency response \cite{chung2023into}. Furthermore, human–robot collaboration has also been applied in social navigation within public spaces \cite{singamaneni2024survey}. However, little research has explored human–robot collaboration for joint optimization of heterogeneous urban services, which is the focus of our work.

\section{Conclusion}
This paper presents \M, a two-layer human-robot collaboration framework designed for joint optimization of heterogeneous urban services (e.g., crowdsourced delivery and urban sensing). \M comprises two key components: (i) a scalable distributed MapReduce-based K-Submodular maximization module called \UM for real-time order dispatch, and (ii) a deep submodular reward Q-network algorithm named \DM for dynamic urban sensing. These two modules are \textbf{\textit{interconnected}} as the \DM will also provide sensing values for the \UM to optimize and balance real-time delivery and sensing performance.
Our evaluation on real-world datasets from a food delivery platform demonstrates that \M improves sensing coverage by an average of 29.7\% (with 1,000 to 4,000 RVs) and increases courier income by 39.2\% on average, while substantially reducing overdue orders.

\section*{Acknowledgment}
We sincerely thank all anonymous reviewers for their valuable suggestions. This work is partially supported by Florida State University, National Science Foundation under Grant Numbers 2411152, 2441179, 2231523, and 2140346.

\bibliographystyle{IEEEtran}  
\bibliography{sample}



\end{document}